%% file: main.tex
\documentclass[twocolumn]{galbot}

\usepackage{microtype}
\usepackage{graphicx}
\usepackage{subcaption}
\usepackage{booktabs} 
\usepackage{svg}
\usepackage{multirow}
\usepackage{makecell}
\usepackage{pifont}
\usepackage{enumitem}
\usepackage{tabularx}
\usepackage{hyperref}

\usepackage{amsmath}
\usepackage{amssymb}
\usepackage{mathtools}
\usepackage{amsthm}
\usepackage[most]{tcolorbox}
\usepackage[table]{xcolor}

\theoremstyle{plain}

\theoremstyle{definition}

\theoremstyle{remark}

\newcommand{\method}{VLA-JEPA}
\newcounter{question}
\newcommand{\question}{%
    \stepcounter{question}%
    \noindent\textbf{Q\thequestion:~\ignorespaces}%
}

\definecolor{linecolor1}{RGB}{246, 248, 239}
\definecolor{linecolor2}{RGB}{230, 234, 217}
\definecolor{linecolor3}{RGB}{211, 222, 190}

\newcommand{\infobox}[1]{
    \vspace{-0.18cm}
    \begin{tcolorbox}[
        colback=white!90!gray,     
        colframe=teal!60!black,   
        arc=5pt,                   
        boxsep=5pt,                 
        left=5pt,                  
        right=10pt,                 
        top=2pt,                   
        bottom=3pt,                
        boxrule=0.8pt,              
        drop shadow=gray!50!white, 
        enhanced jigsaw             
    ]
    \vspace{-0.1cm}
         \textit{#1}
    \vspace{-0.2cm}
    \end{tcolorbox}
    \vspace{-0.15cm}
}

\usepackage[textsize=tiny]{todonotes}

\title{VLA-JEPA: Enhancing Vision-Language-Action Model with Latent World Model}

\author[1, 2, *]{Jingwen Sun}
\author[3, 5*]{Wenyao Zhang}
\author[4]{Zekun Qi}
\author[2, 6]{Shaojie Ren}
\author[2, 7]{Zezhi Liu}
\author[1]{Hanxin Zhu}
\author[1]{Guangzhong Sun}
\author[2, 5, \dagger]{Xin Jin}
\author[1, 2, \dagger]{Zhibo Chen}

\affiliation[1]{University of Science and Technology of China}
\affiliation[2]{Zhongguancun Academy, Beijing, China}
\affiliation[3]{Shanghai Jiao Tong University}
\affiliation[4]{Tsinghua University}
\affiliation[5]{Eastern Institute of Technology, Ningbo}
\affiliation[6]{University of Chinese Academy of Sciences}
\affiliation[7]{Nankai University}

\contribution[*]{Equal Contribution}
\contribution[\dagger]{Corresponding author}

\date{\today} 
\code{\url{https://github.com/ginwind/VLA-JEPA/}}
\page{\url{https://ginwind.github.io/VLA-JEPA/}}
\huggingface{\url{https://huggingface.co/ginwind/VLA-JEPA/}}

\abstract{
Pretraining Vision-Language-Action (VLA) policies on internet-scale video is appealing, yet current latent-action objectives often learn the wrong thing: they remain anchored to pixel variation rather than action-relevant state transitions, making them vulnerable to appearance bias, nuisance motion, and information leakage. We introduce \method{}, a JEPA-style pretraining framework that sidesteps these pitfalls by design. The key idea is \emph{leakage-free state prediction}: a target encoder produces latent representations from future frames, while the student pathway sees only the current observation---future information is used solely as supervision targets, never as input. By predicting in latent space rather than pixel space, \method{} learns dynamics abstractions that are robust to camera motion and irrelevant background changes. This yields a simple two-stage recipe---JEPA pretraining followed by action-head fine-tuning---without the multi-stage complexity of prior latent-action pipelines. Experiments on LIBERO, LIBERO-Plus, SimplerEnv and real-world manipulation tasks show that \method{} achieves consistent gains in generalization and robustness over existing methods.
}

\begin{document}
\maketitle
\pagestyle{empty}

\section{Introduction}
\label{sec:intro}
\input{1_Intro}

\section{Related Works}
\input{2_RW}

\section{Methodology}
\input{3_Methodology}

\section{Experiments}
\input{4_Experiments}

\section{Conclusion}
This paper proposes \method{}, a unified pretraining framework that learns latent actions from both human and robot videos via latent world modeling. Extensive experiments and analyses reveal that existing latent action pretraining methods suffer from information leakage and representation degeneration, which hinder the learning of meaningful temporal dynamics. In contrast, \method{} effectively mitigates these issues and enables the model to capture genuine inter-frame dynamics. As a result, \method{} achieves competitive performance in both simulation benchmarks and real-world robotic experiments.
We further believe that the human-video pretraining paradigm introduced by \method{} is highly scalable, and can be naturally extended by incorporating robot data and text-based reasoning data, thereby further improving the generalization and robustness of VLA models.

\section*{Acknowledgement}
This work is supported by the Zhongguancun Academy, (Grant No.s C20250302)

\bibliographystyle{assets/plainnat}
\bibliography{main}

\newpage
\appendix
\onecolumn
\input{Appendix}

\end{document}

%% file: 1_Intro.tex
Learning visuomotor policies from internet-scale video has become an increasingly attractive route for robot learning~\cite{grauman2022ego4d,goyal2017something}. 
Compared to robot interaction data, which are costly and narrow in coverage~\cite{chi2024umi,lin2024data}, unlabeled videos are abundant and diverse, offering rich demonstrations of temporally extended change. 
This has motivated a growing body of \emph{latent-action pretraining} for Vision--Language--Action (VLA)~\cite{RT123,kim2024openvla,24pi0,25pi0.5}: rather than learning actions only from scarce control trajectories, they first learn representations and transition structure from video, then adapt them to downstream control~\cite{ye2024lapa,bu2025univla,moto24}.

However, we argue that today's ``latent action from video'' objectives often do not learn what we actually need for control. 
For an embodied agent, the most useful notion of ``action'' is not a compact descriptor of pixel differences; it is a variable that captures how the underlying state will evolve under interaction, i.e., \emph{action-relevant state transition semantics}. 
When pretraining is misaligned with this goal, the downstream policy inherits representations that are temporally predictive yet weakly tied to controllable structure---leading to brittle behavior, poor transfer, and inefficient fine-tuning.

\vspace{5pt}
\infobox{Why latent-action pretraining often drifts away from action semantics?}
\vspace{6pt}

We identify four failure modes that repeatedly appear in latent-action pretraining pipelines built on unlabeled video:

\begin{enumerate}[leftmargin=*,itemsep=0pt,topsep=0pt]
    \item \textbf{Pixel-level objectives bias representations toward appearance, not action.}
    A common strategy is to use ``the future'' as supervision---either by predicting future pixels directly or by compressing frame-to-frame changes into a latent variable interpreted as an action~\cite{ye2024lapa,bu2025univla,zhang2025dreamvla}. 
    Even with compression mechanisms such as VQ-VAE~\cite{van2017vqvae}, the supervision signal is frequently dominated by what changes visually: texture, illumination, background clutter, and viewpoint. 
    These factors are high-variance but low-control; they are easy to predict yet only weakly connected to the controllable degrees of freedom that a policy must master.
    
    \item \textbf{Real-world videos amplify noisy motion.}
    This mismatch becomes especially pronounced on human videos and in-the-wild footage~\cite{grauman2022ego4d,vpp2024hu}, where camera motion and non-causal background changes can be stronger than interaction-induced state changes. 
    Frame-difference-based latent-action objectives are then incentivized to encode these dominant signals, turning the latent action into a delta-frame encoder of nuisance motion rather than a representation of meaningful transition dynamics.
    
    \item \textbf{Information leakage makes ``latent action'' collapse into a shortcut.}
    Several latent-action pipelines model transitions by feeding both the current observation and future observation into the same module, or by allowing future context to influence the learned action variable during training~\cite{ye2024lapa,bu2025univla}. 
    This design creates an easy shortcut: the latent action can simply encode the future itself instead of capturing how state transitions should be explained~\cite{zhang2025latent}. 
    The resulting ``action'' becomes semantically empty---useful for matching training losses, but not a meaningful factor for control.
    
    \item \textbf{Multi-stage training pipelines are complex and fragile.}
    To stabilize training and mitigate the above issues, many approaches rely on three-stage (or more) procedures: representation pretraining, latent-action learning/alignment, then policy learning~\cite{ye2024lapa,chen2025villa,fan2025xr}. 
    These pipelines increase engineering complexity, introduce stage-wise inconsistencies, and make it harder to train and evaluate methods cleanly.
\end{enumerate}

\begin{figure}[t!]
    \centering
    \includegraphics[width=1\linewidth]{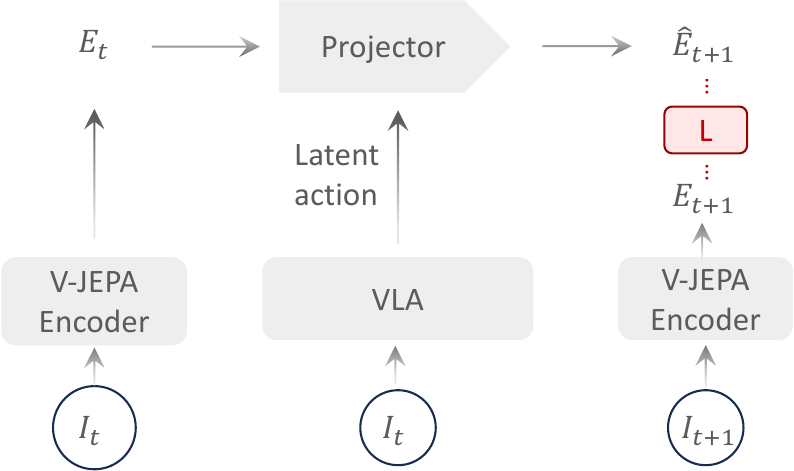}
    \caption{\centering{VLA-JEPA model architecture}}
    \label{fig:intro}
    \vspace{-5pt}
\end{figure}

Together, these issues point to a single underlying problem: many latent-action objectives remain implicitly anchored to pixel variation, and thus learn dynamics that are predictive but not necessarily action-centric~\cite{zhang2025latent}.
For embodied control, we want a \emph{value-bearing latent state} that discards nuisance appearance while preserving factors governing state evolution under interaction~\cite{vpp2024hu,defi,wen2024vidman,jia2025video2act}. 
This motivates a key principle: \emph{predict future latent states that reflect action-relevant transition structure, while preventing future information from leaking into the predictor.}

This aligns naturally with JEPA (Joint-Embedding Predictive Architectures)~\cite{bardes2023vjepa1,assran2025vjepa2,assran2023I-JEPA, destrade2025valueguidedactionplanningjepa}, which replaces pixel reconstruction with latent-space alignment. 
By predicting representations rather than pixels, JEPA is inherently robust to low-level noise and encourages semantic abstraction~\cite{garrido2026learning}.

Motivated by this perspective, we introduce \method{}, a JEPA-style pretraining framework tailored to VLA policies. 
As shown in Figure~\ref{fig:intro}, the key design is \emph{leakage-free state prediction}: during pretraining, a target video encoder produces latent targets from future context (e.g., a short clip), while the latent action pathway receives only the current observation through a VLM backbone. 
A predictor maps the history latent states and the latent action representations to future latent states, which is trained as a latent world model using a JEPA alignment loss. Crucially, future frames are never provided as inputs to the VLM backbone; instead, they are used solely to construct the training targets, eliminating the shortcut that causes latent-action collapse.
This yields two benefits: (1) semantic robustness to camera motion and background changes, since supervision operates in latent space rather than pixel space; (2) a streamlined two-stage pipeline—JEPA pretraining followed by action-head fine-tuning—without introducing auxiliary modules or redefining the learned representations.
This work makes three contributions:

\begin{itemize}[leftmargin=*,itemsep=0pt,topsep=0pt]
    \item \textbf{Analysis of latent-action pretraining pitfalls.} We analyze why many future-supervision objectives (including compressed frame-difference latent actions) remain pixel-tethered, becoming biased toward appearance, vulnerable to nuisance motion in real-world videos, and prone to information leakage when future context enters the learner.
    
    \item \textbf{VLA-JEPA: leakage-free, state-level JEPA pretraining.} We propose a JEPA-style latent predictive alignment scheme that learns action-relevant transition semantics by predicting and aligning future latent states---without pixel reconstruction, information leakage and only one-stage pretraining pipeline.
    
    \item \textbf{Improved and robustness with a simpler workflow.} Across embodied control benchmarks (LIBERO, LIBERO-Plus, SimplerEnv) and real-world settings, \method{} yields consistent gains in robustness, and generalization, while simplifying training relative to prior multi-stage latent-action pipelines.
\end{itemize}

%% file: 2_RW.tex
\subsection{Vision-Language-Action Models}
With the rapid advancement of Large Language Models (LLMs)~\citep{llava23, karamcheti2024prismatic, beyer2024paligemma,DreamLLM23,yang2025qwen3,ShapeLLM24} and large-scale robot datasets~\citep{o2023open, ebert2021bridge, khazatsky2024droid, deng2025graspvla}, Vision-Language-Action (VLA) models have become a dominant paradigm in robot learning~\cite{kim2024openvla,24pi0,xu2025seeing,Chen2025SimpleVLA-RL}. 
The RT series~\citep{RT123, RT223, RTH24} pioneered fine-tuning multimodal LLMs on robot demonstrations, and subsequent works further improved manipulation and navigation performance~\citep{kim2024openvla, 24pi0, qu25spatialvla, liang2025discrete,zhang2024navid,zhang2024uni,pretrainingdisentangled,wu2026pragmatic}.
However, most VLA methods rely heavily on large-scale action-labeled robot data, which is costly and difficult to scale~\cite{wang2025vla}. 
To reduce dependence on explicit action supervision, recent works introduce multimodal Chain-of-Thought signals, such as hierarchical planning~\citep{RTH24,lin2025onetwovla,sofar25,zhou2025chatvla,ji2025robobrain}, subgoal or rollout prediction~\citep{seer24, cotvla25, 3dvla, wang2025unified,lin2025hif,chen2025unified,zhang2025future,liu2025mla}, object-centric conditioning~\citep{deng2025graspvla,25pi0.5,ranasinghe2025pixel,zhong2025flowvla,upvla25,lv2025f1,cai2026internvla} and latent future embeddings or actions~\citep{go125bu,dywa25,zhang2025dreamvla,liu2026last,zhong2026acot}.
Nevertheless, these approaches suffer from relying on action-labeled data. 
In contrast, ~\method{} learns action-centric representations via latent predictive alignment, avoiding explicit future reconstruction and reducing the need for large-scale action supervision.

\subsection{Latent action learning for robotics}
To leverage large-scale videos without action labels, ILPO~\cite{edwards2019imitating}, LAPO~\cite{23lapo} and Genie~\cite{bruce2024genie} propose using latent action for video games.
For robot learning, LAPA~\cite{ye2024lapa}, IGOR~\cite{chen2024igor}, UniVLA~\cite{bu2025univla}, MotoGPT~\cite{moto24}, Adaworld~\cite{gaoa24daworld}, CoMo~\cite{yang2025como} and StaMo~\cite{liu2025stamo} extract discrete or continuous motion tokens from frame transitions, and pretrain VLA to predict these latent actions before mapping them to real robot controls. 
To align the latent action to the real action space, villa-x~\cite{chen2025villa}, XR-1~\cite{fan2025xr}, CLAP~\cite{zhang2026clapcontrastivelatentaction} and VITA~\cite{ma2025unifying} propose that extract latent action from both robot and human video and use a unified codebook.
However, since latent actions are often learned directly from adjacent frames, models may exploit pixel-level shortcuts and encode future-frame leakage.
Although some methods attempt to mitigate the above issues by constraining the latent action space—through optical-flow~\cite{bu2025laof,bi2025motus}, or object-centric constraints~\cite{Alexander2025Object}, among others~\cite{Alexander2025LALRS,Alexander2026VLM}—such constraints inevitably bias the learned latent actions toward aligning with hand-crafted, visually defined priors rather than the true underlying controllable factors. Consequently, when such priors become mismatched in novel environments, the learned latent actions tend to fail systematically, and the latent spaces may align more with visual deltas than with actionable control signals, necessitating multi-stage training pipelines and additional alignment mechanisms~\cite{zhang2025latent}.
On the contrary, our proposed ~\method{} learns action-relevant representations without relying on delta frame info extraction, avoiding leakage and pixel shortcuts while enabling single-stage end-to-end pretraining.

\begin{figure*}[t!]
    \centering
    \includegraphics[width=0.98\linewidth]{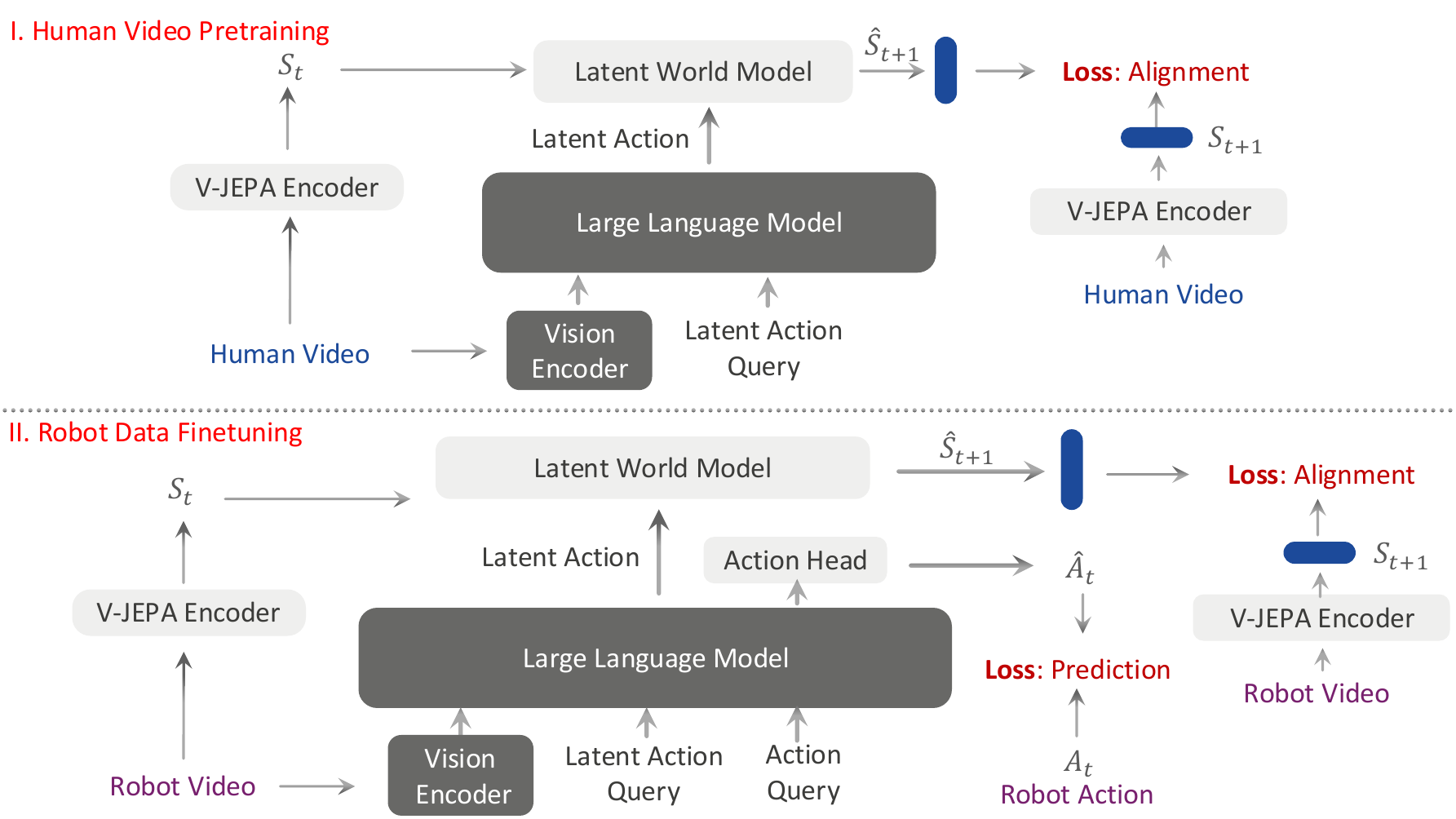}
    \caption{\method{} supports cross-domain training on both human videos and robot data, where human videos are trained using an alignment loss under the latent world modeling objective, while robot data are trained with a joint objective consisting of an alignment loss and a robot action prediction loss.}
    \label{fig:jevla}
\end{figure*}

%% file: 3_Methodology.tex
To address the limitations of existing approaches, we introduce \method{}, a unified framework that enables joint pre-training on both action-free and action-labeled data. For action-free human videos, \method{} extracts latent actions from vision-language prior representations by optimizing a world-model-based state transition objective. Building upon this foundation, we further integrate a flow-matching-based action generator for robot demonstrations to support precise end-effector trajectory generation. During fine-tuning, \method{} enables end-to-end fusion of the two objectives, allowing the learned state-transition dynamics to be effectively leveraged for downstream robotic control.

\subsection{Model Backbone}
As shown in Figure~\ref{fig:jevla}, we adopt Qwen3-VL~\cite{Bai25Qwen3VL} as the core large vision-language model (VLM) in our framework. This VLM is built upon the Qwen3~\cite{yang2025qwen3} and uses SigLIP-2~\cite{tschannen2025siglip2} as its vision encoder. 
It is well established that the world knowledge acquired by VLMs during large-scale pretraining, including image understanding and key object detection, can be transferred to robot control tasks. To distill continuous latent action representations from the VLM, which encode state transition information for world modeling, we introduce a set of learnable tokens that denoted as $\langle latent_i \rangle$ and $\langle action \rangle$, where i denotes the time step. For instance, $\langle latent_0 \rangle$ represents the state transition between $s_0$ and $s_1$. To model state transitions in continuous videos, we encode pixel-level human videos into time-step–aware feature sequences using an encoder architecture trained with V-JEPA2~\cite{assran2025vjepa2}, and employ a time-causal attention mechanism to capture the correlations between the feature sequences and the latent actions.

\subsection{Learning from Human Videos}
To enable VLM to learn from human demonstration videos, we design a training framework that explicitly injects environment dynamics into the latent action tokens. Specifically, we consider a human video dataset $D=\{(O_0,O_1,...,O_v,\ell)\}$, where $\ell$ denotes the associated language instruction, and each $O_v$ represents a video captured from viewpoint $v$. Formally, $O_v=(I_{v,t_0},I_{v,t_1},...,I_{v,t_n})$, where $I_{v,t_i}$ denotes the video frame at time step $t_i$ from view $v$, and $n$ is the total number of frames within a video. 

\textbf{World State Encoder.} Unlike traditional single-view video representation approaches, we encode diverse observations from the same episode into a unified world state representation through a world state encoder. Specifically, we adopt a self-supervised V-JEPA2 encoder as the single-view video state representation and integrate representations from multiple viewpoints via a concatenation operator. The encoding process for each view and the subsequent aggregation across views are formulated as follows:

\begin{equation}
\label{eq:world_state_encoding}
\begin{aligned}
    s_{t_i}=\Vert_v F(I_{v,t_i})
\end{aligned}
\end{equation}
where $F(\cdot)$ is the single-view video encoder (e.g., V-JEPA2), and $\|$ denotes the vector concatenation operator. The resulting $s_{t_i}$ is the unified world-state representation at timestamp $t_i$.

\textbf{Latent Action Pretraining through World Modeling.} To encourage the learnable latent action tokens to capture state transition dynamics, we introduce a world state prediction objective based on an auto-regressive transformer-based world model.

Formally, the VLM takes as input the multi-view observations at the initial time step $t_0$, together with the language instruction $\ell$. Conditioned on these inputs, the VLM maps a set of special learnable tokens $\langle \text{latent}_i \rangle$ into latent representations that summarize the underlying world dynamics:
\begin{equation}
\label{eq:VLM_encoding}
z_{t_i}
=
p_\theta^{VLM}\!\left(
\langle \text{latent}_i \rangle
\;\middle|\;
\{I_{j,t_0}\}_{j=0}^{v}, \ell
\right),
\end{equation}
where $z_{t_i}$ denotes the latent representation associated with the $i$-th latent action token at time step $t_i$.

Subsequently, the unified representation $z_{t_i}$ is used to condition the world model, providing additional context for state prediction. Formally, given the sequence of encoded world states $s_{t_{0:i}}$ and the corresponding conditioning variables $z_{t_{0:i}}$, the world model predicts the next chunk of states as:
\begin{equation}
\label{eq:wm_prediction}
    \hat{s_{t_{1:i+1}}} = p_\theta^{WM}(s_{t_{0:i}}, z_{t_{0:i}}),
\end{equation}
where $\hat{s_{t_{1:i+1}}}$ is the predicted world state chunk over $[t_1, t_{i+1}]$.

In practice, each special latent token $\langle \text{latent}_i \rangle$ is replicated $K$ times in the input sequence to enable variable-length latent action encoding, where $K$ is a tunable hyperparameter. The world model employs a time-causal attention mechanism. Within each time step, all latent action tokens and world state tokens attend to each other via bidirectional full attention. Across different time steps, attention is strictly causal: tokens at time step $t$ are allowed to attend only to tokens from time steps up to and including $t$, while attention to future time steps is masked out.


From the perspective of joint-embedding predictive architecture~\citep{bardes2023vjepa1,assran2025vjepa2}, our training objective can be interpreted as maximizing the \textit{evidence lower bound} (ELBO) of the predictive log-likelihood in semantic space. Specifically, given the frozen V-JEPA2 encoder $F(\cdot)$ that produces target world states $s_{t_i}$, and the world model $p_\theta^{WM}$ conditioned on $z_{t_i}$ that predicts $\hat{s}_{t_i}$, the objective can be written as:
\begin{align}\label{eq:VLAJEPA_ELBO}
\log p(s_{t_{1:T}} \mid z_{t_{0:T-1}})
\geq{} & \sum_{k=1}^{T}
\mathbb{E}_{s_{t_k} \sim F(\cdot)}
\big[\log p_\theta(\hat{s}_{t_k} \mid s_{t_k})\big] \notag \\
& - D_{\mathrm{KL}}
\big[F(\cdot) \| p_\theta^{WM}\big],
\end{align}
where $F(\cdot)$ serves as the frozen target encoder (with stop-gradient), and $p_\theta^{WM}$ is the online predictor. In practice, since $F(\cdot)$ produces deterministic embeddings, the KL term vanishes, and the ELBO reduces to a reconstruction loss in the latent space.


Finally, we optimize the combined WM and VLM using a teacher-forcing objective. This enables the unified representation $z_{t_i}$ informed by the world knowledge encoded in the VLM, in order to effectively characterize world state transitions. The world modeling loss is defined as:
\begin{equation}
\label{eq:world_loss}
    \mathcal{L}_{\mathrm{WM}} = \sum_{k=1}^{T} \mathbb{E}_{s_{t_k} \sim F(\cdot)}(\hat{s}_{t_k} - s_{t_k}),
\end{equation}
where $s_{t_k}$ denotes the ground-truth world state at time $t_k$, $\hat{s}_{t_k}$ is the corresponding prediction, and $T$ is the video prediction horizon.

\begin{figure*}[t!]
    \centering
    \vspace{-10pt}
    \includegraphics[width=1\linewidth]{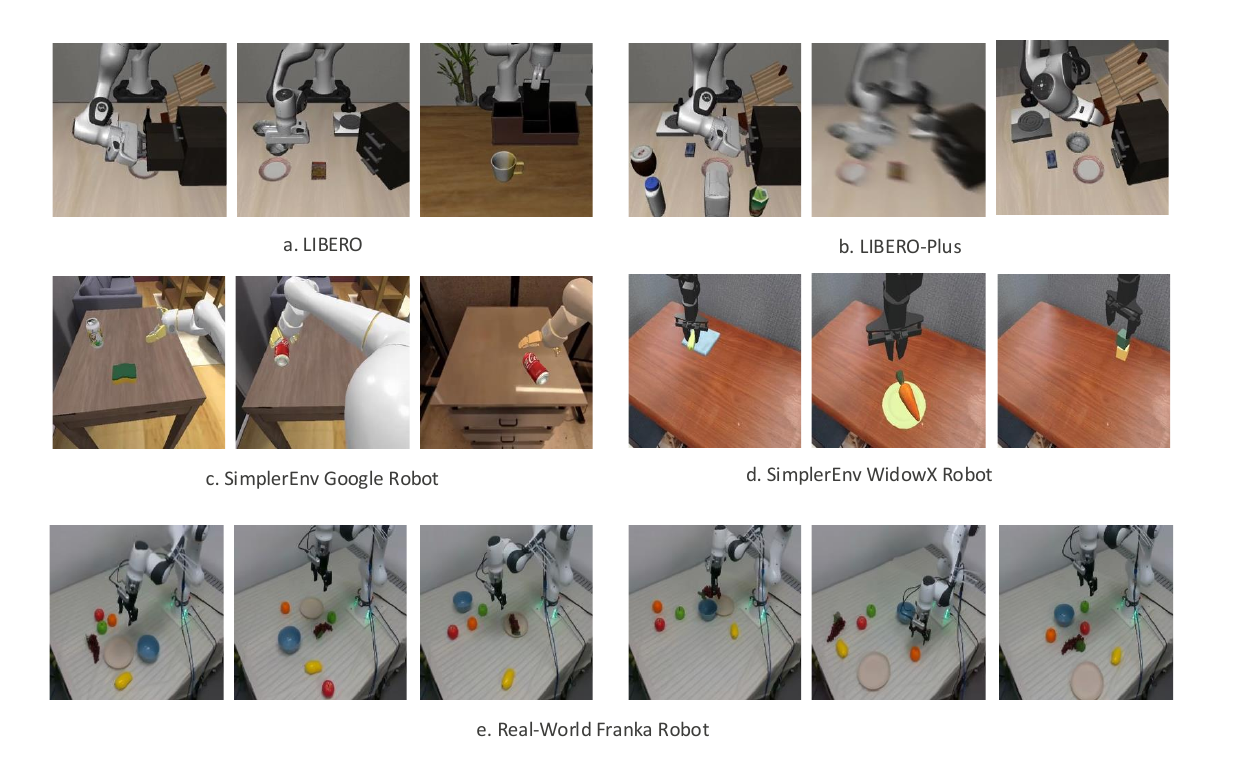}
    \vspace{-24pt}
    \caption{Experiments setup on LIBERO, LIBERO-Plus, SimplerEnv and real-world Franka robot. We evaluate \method{} on 3 simulation benchmarks and 1 real-world environment.}
    \label{fig:experiments_set_up}
    \vspace{5pt}
\end{figure*}

\subsection{Action Prediction with Joint Optimization Objectives}
\textbf{Action Token Conditioning.} To leverage the latent action representations learned from video data for guiding action prediction, we design joint optimization objectives. Specifically, for multi-view RGB videos in the robot dataset, we adopt the same training objective as in Equation~\ref{eq:world_loss} to fine-tune the latent action representations within the robot data domain. For practical action prediction, we intend the latent action to serve as a conditioning signal for embodied action generation, analogous to the roles of the initial image observation $I_{v, t_0}$ and the language instruction. Therefore, we append a set of learnable embodied action tokens $\langle \text{action} \rangle$ after the latent action tokens. Leveraging the causal attention mechanism of the VLM, the model captures the dependencies among $\langle \text{action} \rangle$, the latent action tokens, the initial visual observation, and the language instruction. 

Formally, given the visual tokens $\{I_{i,t_0}\}_{i=0}^{v}$ at the initial time step $t_0$, the language instruction $l$, and a sequence of latent action tokens $\langle \text{latent}_i \rangle$, we obtain a global action-conditioning representation
\begin{equation}
\label{eq:embodied_action_conditioning}
z_{a}
=
p_\theta^{VLM}\!\left(
\langle \text{action} \rangle
\;\middle|\;
\{I_{i,t_0}\}_{i=0}^{v}, \ell, \langle \text{latent}_i \rangle
\right),
\end{equation}
where $z_a$ serves as an additional conditioning signal for the flow-matching-based action head.

\textbf{Conditional Flow-Matching Action Head.} We adopt conditional flow matching to model a distribution over continuous action trajectories. Specifically, let $a_{0:H}$ denote the ground-truth action sequence over a horizon $H$, and $\hat{a}_{0:H}$ denote the predicted action sequence generated by the learned flow. Following standard flow-matching formulations, we define a time-dependent interpolation
\begin{equation}
a_t = (1 - t)\,\epsilon + t\,a_{0:H}, \quad t \sim \mathcal{U}(0,1),
\end{equation}
where $\epsilon \sim \mathcal{N}(0, I)$ is Gaussian noise. The action head parameterizes a vector field $v_\theta(a_t, t \mid z_a)$, conditioned on $z_a$, which is trained to match the ground-truth conditional flow.

The flow-matching objective is given by
\begin{equation}
\mathcal{L}_{\mathrm{FM}}
= \mathbb{E}_{a_{0:H},\, \epsilon,\, t}
\Big[
\big\|
v_\theta(a_t, t \mid z_a)
- (a_{0:H} - \epsilon)
\big\|_2^2
\Big],
\end{equation}
where $v_\theta(\cdot)$ denotes the predicted velocity field, $(a_{0:H} - \epsilon)$ is the target velocity induced by the linear interpolation, and $\|\cdot\|_2$ is the $\ell_2$ norm.

At inference time, the learned vector field is integrated from noise to data space to obtain the predicted action trajectory $\hat{a}_{0:H}$, conditioned on the action tokens $z_a$.

In summary, the overall training objective for action-labeled robot data is as follows:
\begin{equation}
\label{eq:cotrain_loss}
\mathcal{L} = \mathcal{L}_{\mathrm{FM}} + \beta\mathcal{L}_{\mathrm{WM}}, 
\end{equation}
where $\beta$ is a tunable hyperparameter.

%% file: 4_Experiments.tex
\begin{table*}[!htbp]
\caption{Comparison on the LIBERO benchmark. We report the task success rate for each suite and the average across all tasks. Each task is evaluated over 50 episodes (500 episodes per suite), and success rates are computed over these trials. \textbf{Bold} denotes the best performance, and \underline{italics} denotes the second best.}
  \label{table:libero_comparison}
  \begin{center}
    \begin{small}
        \setlength{\tabcolsep}{10pt}
        \begin{tabular}{lccccc}
        \toprule
        \multirow{2}{*}{Method} & \multicolumn{5}{c}{LIBERO}\\
        \cmidrule(lr){2-6}
        & Spatial & Object & Goal & LIBERO-10 & Avg \\
        \midrule
        LAPA~\cite{ye2024lapa} & 73.8 & 74.6 & 58.8 & 55.4 & 65.7 \\
        UniVLA~\cite{bu2025univla} & 96.5 & 96.8 & 95.6 & 92.0 & 95.2 \\
        OpenVLA-OFT~\cite{kim2025fine} & \cellcolor{linecolor1}{\underline{97.6}} & 98.4 & \cellcolor{linecolor1}{\underline{97.9}} & \cellcolor{linecolor1}{\underline{94.5}} & \cellcolor{linecolor1}{\underline{97.1}} \\
        $\pi_0$~\cite{24pi0} & 96.8 & \cellcolor{linecolor1}{\underline{98.8}} & 95.8 & 85.2 & 94.2 \\
        $\pi_0 \texttt{-Fast} $~\cite{pertsch2025fast} & 96.4 & 96.8 & 88.6 & 60.2 & 85.5 \\
        CoT-VLA~\cite{cotvla25} & 87.5 & 91.6 & 87.6 & 69.0 & 81.1 \\
        WorldVLA~\cite{cen2025worldvla} & 87.6 & 96.2 & 83.4 & 60.0 & 81.8 \\
        villa-X~\cite{chen2025villa} & 97.5 & 97.0 & 91.5 & 74.5 & 90.1 \\
        GR00T N1~\cite{bjorck2025gr00t} & 94.4 & 97.6 & 93.0 & 90.6 & 93.9 \\
        $\pi_{0.5}$~\cite{25pi0.5} & \cellcolor{linecolor2}{\textbf{98.8}} & 98.2 & \cellcolor{linecolor2}{\textbf{98.0}} & 92.4 & 96.9  \\
        \midrule
        \method{} & 96.2 & \cellcolor{linecolor2}{\textbf{99.6}} & 97.2 & \cellcolor{linecolor2}{\textbf{95.8}} & \cellcolor{linecolor2}{\textbf{97.2}} \\
        \hspace*{0.5cm} w/o human videos & 94.8 & \cellcolor{linecolor2}{\textbf{99.6}} & 95.8 & 94.0 & 96.1\\
        \bottomrule
        \end{tabular}
        \vspace{5pt}
    \end{small}
  \end{center}
\end{table*}

To evaluate the generalization and robustness of \method{}, we perform comprehensive simulation experiments utilizing two distinct simulation environments and three public benchmarks, Simpler-Env \& LIBERO, along with real-world experiments covering both in-distribution and out-of-distribution settings, as shown in Figure~\ref{fig:experiments_set_up}. We compare VLA-JEPA against the latest VLA baselines that support pretraining with human videos as well as VLA baselines pretrained solely on robot-collected data. 

\subsection{Implementation Details} 
During the pretraining phase, we train all parameters in the model except for the world state encoder. Specifically, we utilize Equation~\ref{eq:world_loss} as the training objective and pretrain on the large-scale human action dataset Something-Something-v2~\cite{goyal2017something}, which contains 220K human videos. The model supports simultaneous pretraining of latent action using both robot action data and human videos without action labels. We also employ Equation~\ref{eq:cotrain_loss} as the training objective and pretrain on the large-scale action-labeled robot dataset Droid~\cite{khazatsky2024droid}, which consists of 76K high-quality demonstration trajectories. 

During fine-tuning, for both LIBERO and LIBERO-Plus, we use the LIBERO dataset, which includes approximately 2K expert demonstrations collected in a simulation environment, without incorporating the augmented dataset from LIBERO-Plus. For SimplerEnv, we utilize the Fractal dataset and the BridgeV2 dataset for post-training, corresponding to the two robot embodiment types in SimplerEnv. For the real-world experiments, we employ a dataset of 100 demonstrations collected across three distinct tasks for post-training. All experiments are conducted on 8 NVIDIA A100 GPUs. Detailed implementation and protocols are provided in Appendix~\ref{appendix:implementation_details}.

\subsection{Simulation Setup and Baseline}

\textbf{Benchmarks.} We conduct generalization experiments on the LIBERO~\cite{LIBERO23} and SimplerEnv~\cite{simplerenv24} benchmarks. The LIBERO benchmark employs the Franka Emika Panda arm and comprises four task suites designed to facilitate research on lifelong learning in robotic manipulation. SimplerEnv features WidowX and Google Robot setups, offering diverse manipulation scenarios with varying lighting, colors, textures, and robot camera poses, thereby bridging the visual appearance gap between real and simulated environments. 
Furthermore, we conduct robustness experiments on LIBERO-Plus~\cite{fei25libero-plus}, a large-scale benchmark designed to systematically stress-test VLAs through perturbed tasks across seven dimensions. 



The three simulation environments correspond to (i) in-distribution scenarios in which policies, trained using simulated expert data, are validated on in-distribution tasks in simulation (LIBERO), (ii) out-of-distribution scenarios involving a real-to-sim gap, in which policies are trained using real-world data and evaluated in simulation (SimplerEnv), and (iii) out-of-distribution scenarios in which policies, trained using simulated expert data, are validated on out-of-distribution tasks in simulation (LIBERO-Plus).

\textbf{Baselines. }We primarily compare VLA-JEPA with previous latent-action VLAs, future-prediction VLAs, as well as state-of-the-art open-source VLAs including Moto~\cite{moto24}, LAPA~\cite{ye2024lapa}, UniVLA~\cite{bu2025univla}, villa-X~\cite{chen2025villa}, CoT-VLA~\cite{cotvla25}, WorldVLA~\cite{cen2025worldvla}, RoboVLMs~\cite{robovlms24}, GR00T N1~\cite{bjorck2025gr00t}, OpenVLA-OFT~\cite{kim2025fine}, $\pi_0$~\cite{24pi0}, $\pi_0 \texttt{-Fast}$~\cite{pertsch2025fast}, and $\pi_{0.5}$~\cite{25pi0.5}.

\subsection{Simulation Evaluation}
\label{sec:simulation_evaluation}

\textbf{LIBERO.} As presented in Table \ref{table:libero_comparison}, VLA-JEPA achieves state-of-the-art performance on 2 out of the 4 task suites within the LIBERO benchmark, securing the highest average success rate overall. We observe that other top-performing models on LIBERO, such as OpenVLA-OFT and $\pi_{0.5}$, rely on extensive robot datasets for pre-training. In contrast, A achieves better performance using less training data. Furthermore, previous latent-action-based VLAs and VLAs trained on human videos, such as UniVLA, villa-X, LAPA, and CoT-VLA, consistently underperform VLA-JEPA, which corroborates the limitations of such approaches discussed in Section~\ref{sec:intro}. 


\textbf{SimplerEnv.} Table~\ref{table:simplerenv_comparison} reports the results on SimplerEnv. VLA-JEPA achieves the best performance on 2 out of 4 tasks for both the Google Robot and the WidowX Robot. In particular, it attains the highest average success rate on the Google Robot and the second highest average success rate on the WidowX Robot. Since SimplerEnv does not provide expert demonstrations, we observe that the quality of robot data plays a crucial role in performance on this benchmark. For example, LAPA extracts only successful rollouts in SimplerEnv and uses them as expert demonstrations for training, and achieves the second-highest success rate on the WidowX Robot with merely 100 rollouts. This is mainly because training on successful rollouts effectively mitigates the real-to-sim gap. Villa-X, on the other hand, is trained on a large-scale collection of robot data and human videos, achieving the highest average success rate on the WidowX Robot. In comparison, while VLA-JEPA and other methods—such as UniVLA, RoboVLMs, and Moto—all utilize less than 1\% of the training data used by villa-X, VLA-JEPA attains the most competitive experimental results. This demonstrates the superiority of our pretraining method.

\textbf{LIBERO-Plus.}
As shown in Table~\ref{table:libero_plus_comparison}, VLA-JEPA achieves the best performance on 5 out of 7 perturbations in LIBERO-Plus. It demonstrates a substantial advantage over UniVLA, which is trained on human videos, and OpenVLA-OFT, $\pi_0$, and other methods trained on large amounts of robot data.
This indicates that latent actions, learned through our pretraining approach, possess a level of world knowledge representation comparable to that encoded by text data. In particular, we find that VLA-JEPA demonstrates a significant advantage over all baselines under perturbations in Language, Light, Background, and Layout, which verifies that our latent action can effectively handle task-agnostic disturbances, thereby achieving a more robust and generalized policy.

\begin{table*}[!htbp]
  \caption{Comparison on the SimplerEnv benchmark. We report the average success rate for each individual task under the visual matching setting in both SimplerEnv-WidowX Robot and SimplerEnv-Google Robot. ${}^*$ denotes that the method is trained on in-distribution expert demonstrations collected in simulation environment. All other methods are trained on subsets of the OXE dataset.}
  \vspace{-2pt}
  \label{table:simplerenv_comparison}
  \begin{center}
    \begin{small}
        \begin{tabular}{lcccccccccc}
          \toprule
          \multirow{2}{*}{Method} & \multicolumn{5}{c}{Google Robot} & \multicolumn{5}{c}{WidowX Robot} \\
          \cmidrule(lr){2-6} \cmidrule(lr){7-11}
          & Pick & Move & Drawer & Place & Avg & Spoon & Carrot & Block & Eggplant & Avg \\
          \midrule
          $\text{LAPA}^*$~\cite{ye2024lapa} & - & - & - & - & - & \cellcolor{linecolor1}{\underline{70.8}} & 45.8 & \cellcolor{linecolor2}{\textbf{54.2}} & 58.3 & \cellcolor{linecolor2}{\textbf{57.3}} \\
            villa-x~\cite{chen2025villa} & 81.7 & 55.4 & 38.4 & 4.2 &44.9& 48.3 & 24.2 & 19.2 & 71.7 & 40.8\\
          UniVLA~\cite{bu2025univla} & - & - & - & - & - & - & - & - & - & 42.7 \\
          RoboVLMs~\cite{robovlms24} & 77.3 & 61.7 & 43.5 & 24.1 & 51.7 & 45.8 & 20.8 & 4.2 & \cellcolor{linecolor2}{\textbf{79.2}} & 37.5 \\
          GR00T N1~\cite{bjorck2025gr00t} & 0.7 & 1.9 & 2.9 & 0.0 & 1.4 & 1.4 & 0.0 & 0.0 & 13.9 & 3.8 \\
          MoTo~\cite{moto24} & 74.0 & 60.4 & 43.1 & - & - & - & - & - & - & - \\
          OpenVLA-OFT~\cite{kim2025fine} & - & - & - & - & - & 34.2 & 30.0 & \cellcolor{linecolor1}{\underline{30.0}} & \cellcolor{linecolor1}{\underline{72.5}} & 41.8 \\
          $\pi_0$~\cite{24pi0} & 72.7 & 65.3 & 38.3 & - & - & 29.1 & 0 & 16.6 & 62.5 & 40.1  \\
          $\pi_0 \texttt{-Fast} $~\cite{pertsch2025fast} & 75.3 & \cellcolor{linecolor2}{\underline{67.5}} & 42.9 & - & - & 29.1 & 21.9 & 10.8 & 66.7 & \cellcolor{linecolor1}{\underline{48.3}} \\
          \midrule
          \method{} & \cellcolor{linecolor2}{\textbf{88.3}} & 64.1 & \cellcolor{linecolor1}{\underline{59.3}} & \cellcolor{linecolor1}{\underline{49.1}} & \cellcolor{linecolor1}{\underline{65.2}} & \cellcolor{linecolor2}{\textbf{75.0}} & \cellcolor{linecolor2}{\textbf{70.8}} & 12.5 & 70.8 & \cellcolor{linecolor2}{\textbf{57.3}} \\
          \hspace*{0.5cm} w/o human videos & \cellcolor{linecolor1}{\underline{85.3}}& \cellcolor{linecolor1}{\underline{66.7}}& \cellcolor{linecolor2}{\textbf{75.5}}& \cellcolor{linecolor2}{\textbf{86.1}} & \cellcolor{linecolor2}{\textbf{78.4}}  &\cellcolor{linecolor2}{\textbf{75.0}} & \cellcolor{linecolor1}{\underline{54.2}} & 20.8 & \cellcolor{linecolor2}{\textbf{79.2}} & \cellcolor{linecolor2}{\textbf{57.3}}\\
          \bottomrule
        \end{tabular}
        \vspace{8pt}
    \end{small}
  \end{center}
\end{table*}

\begin{table*}[t]
  \caption{Comparison on the LIBERO-Plus benchmark. We report the average success rate across each perturbation dimension, where each perturbation includes the four task suites from the original LIBERO benchmark.}
  \vspace{-2pt}
  \label{table:libero_plus_comparison}
  \begin{center}
    \begin{small}
        \begin{tabular}{lcccccccc}
        \toprule
        \multirow{2}{*}{Method} & \multicolumn{8}{c}{LIBERO-Plus}\\
        \cmidrule(lr){2-9}
        & Camera & Robot & Language & Light &Background &Noise &Layout &Avg\\
        \midrule
        UniVLA~\cite{bu2025univla} & 1.8 & 46.2 & 69.6 & 69.0 &81.0 &21.2 &31.9 &42.9 \\
        OpenVLA-OFT~\cite{kim2025fine} & 56.4 & 31.9 & \cellcolor{linecolor1}{\underline{79.5}} & \cellcolor{linecolor1}{\underline{88.7}} & \cellcolor{linecolor1}{\underline{93.3}} &\cellcolor{linecolor1}{\underline{75.8}} & 74.2 & \cellcolor{linecolor1}{\underline{69.6}} \\
        $\pi_0$~\cite{24pi0} & 13.8 &6.0 &58.8 &85.0 &81.4 &\cellcolor{linecolor2}{\textbf{79.0}}& 68.9 &53.6 \\
        $\pi_0 \texttt{-Fast} $~\cite{pertsch2025fast} & \cellcolor{linecolor2}{\textbf{65.1}} &21.6 &61.0 &73.2 &73.2 &74.4 &68.8 &61.6\\
        WorldVLA~\cite{cen2025worldvla} & 0.1& 27.9& 41.6& 43.7 &17.1& 10.9& 38.0& 25.0 \\
        \midrule
        \method{} & \cellcolor{linecolor1}{\underline{63.3}} & \cellcolor{linecolor2}{\textbf{67.1}} & \cellcolor{linecolor2}{\textbf{85.4}} & \cellcolor{linecolor2}{\textbf{95.6}} & \cellcolor{linecolor2}{\textbf{93.6}} & 66.3 & \cellcolor{linecolor2}{\textbf{85.1}} & \cellcolor{linecolor2}{\textbf{79.5}} \\
        \hspace*{0.5cm} w/o human videos & 40.3 & \cellcolor{linecolor1}{\underline{55.7}} & 72.9 & 88.2 & 70.5 & 38.2 & \cellcolor{linecolor1}{\underline{74.6}} &   62.9\\
        \bottomrule
        \end{tabular}
        \vspace{6pt}
    \end{small}
  \end{center}
\end{table*}

\subsection{Real-world Experiments}
For real-world experiments, we designed table-top manipulation tasks using a Franka Research 3 arm equipped with a Robotiq 2F-85 gripper. We collected 100 human demonstration trajectories for training, which include 3 picking and placing tasks. In alignment with the simulation experiments and extending beyond in-distribution task evaluation, we introduce two OOD experimental protocols to rigorously assess the model's generalization capability and robustness for physical deployment. The first OOD protocol involves executing tasks not present in the training data, validating the model's ability to acquire and transfer fundamental skills. The second protocol requires executing tasks seen during training but with randomized object layouts, thereby emulating the cluttered scenarios typical of real-world table-top manipulation tasks. For fair comparison, we finetune $\pi_0$ and $\pi_{0.5}$ on collected demonstration datasets. The detailed task settings can be found in Appendix~\ref{appendix:real-world}.

\begin{figure}[t!]
    \centering
    \includegraphics[width=1\linewidth]{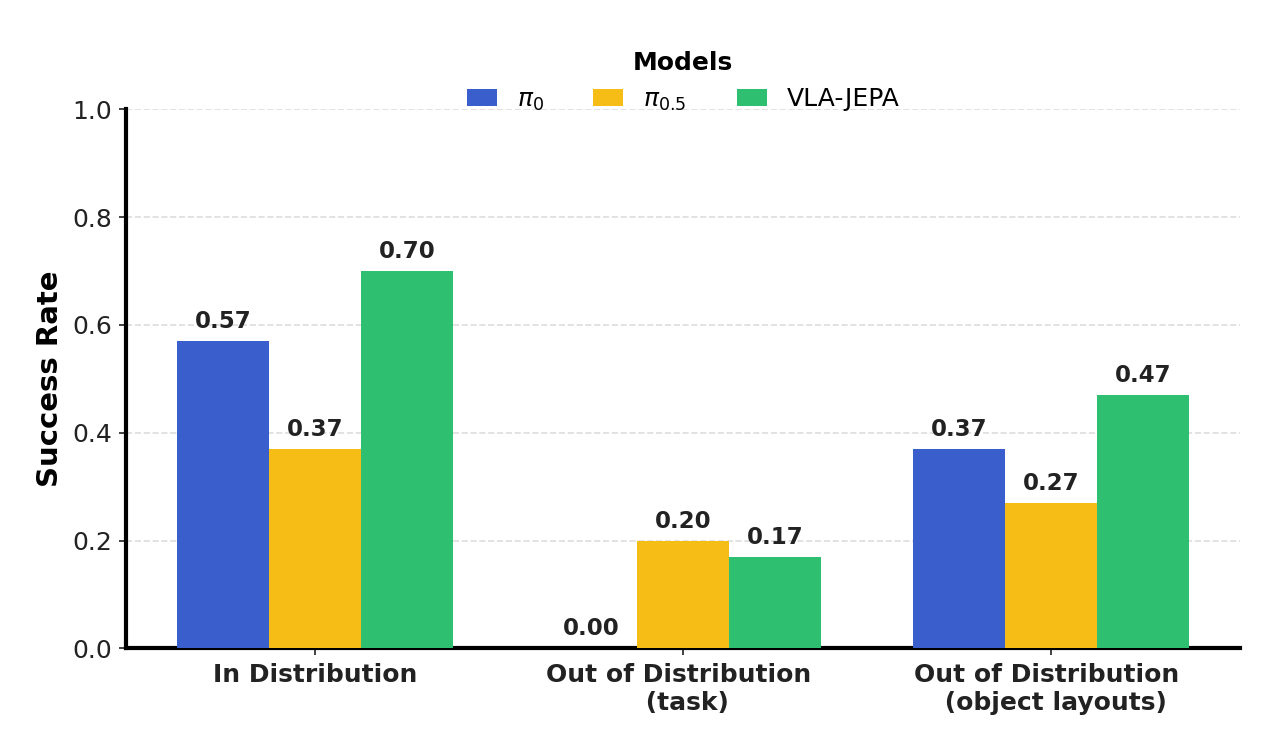}
    \vspace{-20pt}
    \caption{\centering{Real World Experimental Results}}
    \label{fig:real_world_exp}
    \vspace{-8pt}
\end{figure}

As shown in Figure~\ref{fig:real_world_exp}, VLA-JEPA achieves state-of-the-art performance under both the ID and the object layout OOD settings, demonstrating a clear advantage over both $\pi_0$ and $\pi_{0.5}$. In the task OOD setting, it attains the second-best result. During real-world deployment, we observed that the generalization capability of VLA-JEPA is less robust than that of $\pi_{0.5}$, yet it produces execution trajectories that are significantly more stable and reliable. Specifically, $\pi_{0.5}$ more accurately follows instructions to contact the target object compared to VLA-JEPA. However, its position control frequently violates the safety boundaries of the robot arm, leading to execution failures. In contrast, VLA-JEPA, due to its lack of fine-grained reasoning over textual instructions, is prone to grasping objects that do not align with the command. Nevertheless, it rarely breaches the robot arm's safety constraints.

In addition, VLA-JEPA acquires the skill of repeated grasping, i.e., reopening the gripper to attempt another grasp after a failure, which is not observed in $\pi_0$ or $\pi_{0.5}$. We attribute this to the abundance of repeated-grasping knowledge present in human videos, whereas this skill does not require additional physical dynamics knowledge. In contrast, robot data rarely contain demonstrations specifically designed for repeated grasping. A corresponding demonstration can be found in the Appendix~\ref{appendix:real-world}.

\subsection{Further Analysis and Ablation Study}
In this section, we investigate the following questions under the multi-view setting to thoroughly evaluate the ability of our model:

\question \textbf{What impact does human video have?}
Tables~\ref{table:libero_comparison},~\ref{table:simplerenv_comparison},~\ref{table:libero_plus_comparison} present a comparison of VLA-JEPA pretrained with human videos versus without such pretraining, demonstrating that ablating the human video component from the pre-training data does not lead to a significant performance drop across LIBERO and SimplerEnv benchmarks. Interestingly, even pretraining without human videos can lead to higher success rates on SimplerEnv. This indicates that, for ID scenario and real-to-sim gap OOD scenario, high-quality expert demonstrations are more critical for ensuring model performance compared to human videos.

On the LIBERO-Plus benchmark, human demonstration videos provide substantial performance gains. Our analysis suggests that this is because human videos lack effective information about action trajectories, preventing VLA models from directly learning the physical dynamics of robotic actions. Instead, the primary benefit lies in enhancing the robustness and stability of the model’s pre-existing skills, such as repeated grasping. This process closely resembles human skill acquisition from observation: when a person learns a skill by watching a video, the first attempt is unlikely to succeed, and proficiency is achieved only after repeated trials that establish the correspondence between the observed video knowledge and the underlying physical dynamics, ultimately leading to a stable and reliable policy.

\begin{figure}[t!]
    \centering
    \includegraphics[width=1\linewidth]{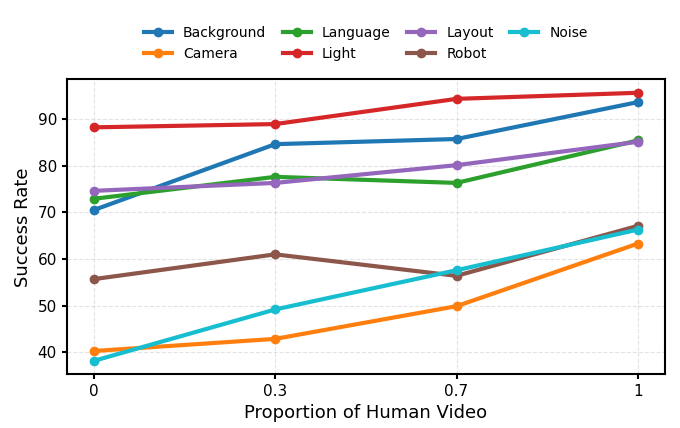}
    \vspace{-15pt}
    \caption{Effect of the proportion of human video data in pre-training on success rates across different perturbation dimensions on the LIBERO-Plus benchmark.}
    \label{fig:human_video_proportion}
\end{figure}

Figure~\ref{fig:human_video_proportion} illustrates how the success rates across different perturbation dimensions on the LIBERO-Plus benchmark vary as the proportion of human videos in the pre-training data increases. These results further corroborate our hypothesis that human video data primarily enhances the robustness and stability of the VLA model by strengthening its existing skill repertoire, rather than introducing new action execution capabilities. Moreover, as the scale of human video data increases, the robustness of the resulting policy consistently improves.

\question \textbf{What is the impact of unified pretraining?}

According to Section~\ref{sec:simulation_evaluation}, the experimental results demonstrate that our unified pretraining method consistently outperforms the previous two-stage pretraining paradigm. In addition, to more clearly analyze the information captured by latent action tokens, we visualize the attention weights from latent action tokens to image tokens in the internal attention maps of three VLAs—LAPA, UniVLA, and \method{}—under simulated, human-video, and real-world robot image inputs. For a fair comparison, all methods are evaluated using pretrained-only checkpoints, without any fine-tuning on either simulated or real data.

\begin{figure*}[t!]
    \centering
    \includegraphics[width=1\linewidth]{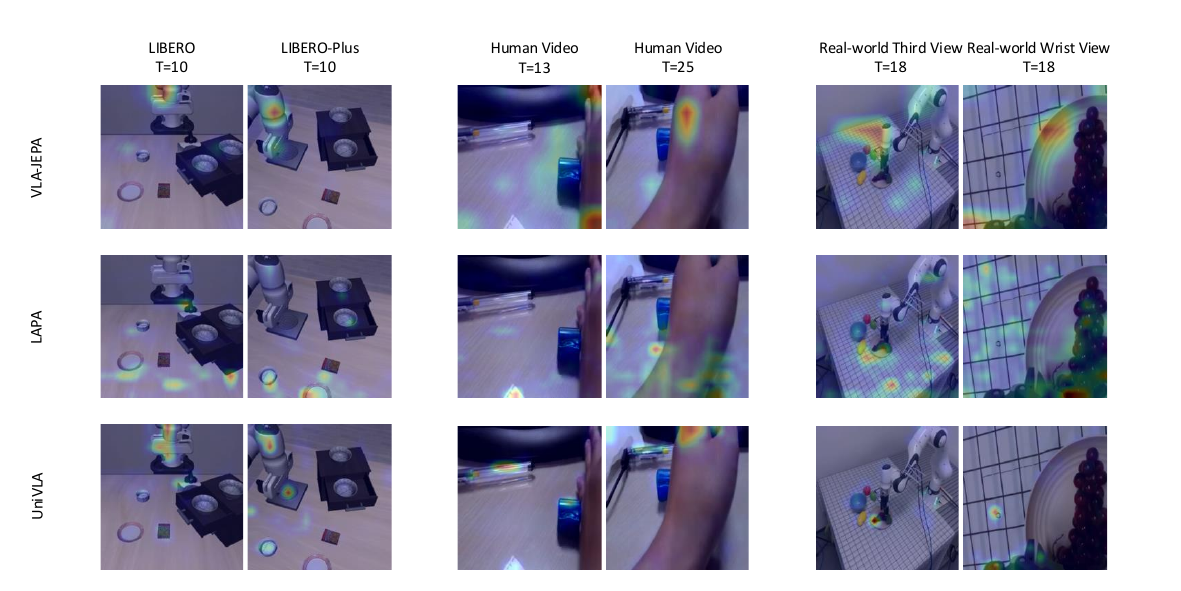}
    \caption{Visualization of the attention weight matrix of latent action tokens attending to image tokens.}
    \label{fig:attention_map}
\end{figure*}

As shown in Figure~\ref{fig:attention_map}, the latent actions of LAPA focus on excessively dense visual information, which results in the inclusion of too many operation-irrelevant details, such as unrelated objects on the desktop. We attribute this to the leakage of information during the pretraining stage, leading to the degradation of latent actions into compressed representations of the target images, a point further analyzed in Section~\ref{sec:intro}. In contrast, UniVLA alleviates this issue through task-relevant textual guidance, yet its overemphasis on semantics causes attention to operation-irrelevant background elements in the images, such as the stationary pen in human videos or the texture of the tablecloth in real-world wrist-view recordings.

By comparison, we focus more precisely on the operation, for instance, the robotic arm, the hand, and the objects to be manipulated. This demonstrates that the unified pretraining approach simplifies the training pipeline while improving training effectiveness by reducing the impact of task-irrelevant information.

\begin{table}[!htbp]
  \vspace{3pt}
  \caption{Comparison of \method{} on the LIBERO benchmark with different numbers of future video horizon.}
  \vspace{-2pt}
  \label{table:libero_comparison_T}
  \begin{center}
    \begin{small}
        \begin{tabular}{cccccc}
        \toprule
        T & Spatial & Object & Goal & 10 & Avg \\
        \midrule
        4 & \textbf{95.0} & 99.2 & 95.8 & 89.0 & 94.8 \\
        8 & 94.8 & \textbf{99.8} & 95.8 & \textbf{94.0} & \textbf{96.1} \\
        16 & 92.8 & 98.8 & \textbf{98.0} & 92.2 & 95.5\\
        \bottomrule
        \end{tabular}
    \end{small}
  \end{center}
\vspace{3pt}
\end{table}

\question \textbf{How does the number of future video horizon affect performance?}

We aim for latent actions to capture the motion dynamics between consecutive frames, such that the number of latent action tokens is always equal to the number of frames minus one. We vary the number of future video horizon $T \in \{4, 8, 16\}$ to investigate the impact of dynamic information on the performance. For a fair comparison, all models are directly fine-tuned on the LIBERO dataset, with all other hyperparameters kept the same in the comparison.

As shown in Table~\ref{table:libero_comparison_T}, the model achieves its best when the video horizon is close to the predefined action horizon, which indicates that latent actions are effective for embodied action generation. When $T$ is too small, the encoded information is insufficient, leading to inferior performance, particularly on long-horizon tasks. In contrast, an excessively large $T$ introduces redundant information. It yields the best results on the goal-oriented task suite with relatively simple objectives, but performs worst on the spatial task suite that requires fine-grained manipulation.


%% file: Appendix.tex
\section{Implementation Details}
\label{appendix:implementation_details}
\subsection{VLA-JEPA Architecture}

\textbf{VLM Backbone.} We use Qwen3-VL-2B as the VLM backbone, which is implemented as a dense Transformer. Its vision encoder consists of a Vision Transformer and 3D convolutional modules. To capture physical dynamics, we adopt a V-JEPA2 encoder checkpoint together with a randomly initialized predictor as the latent world model, whose action inputs are the latent action tokens produced by the VLM. To enable the VLM to output time-aware latent actions and embodied actions, we designed latent action tokens $\langle latent_i \rangle$ and an embodied action token $\langle \text{action} \rangle$, which are incorporated as additional tokens into the vocabulary of Qwen3-VL. When generating the latent action tokens, we repeat the same latent action token $\langle latent_i \rangle$ $K$ times to strengthen the model’s attention to the latent action tokens, where $K = 24 / T$, $T$ denotes the future video horizon, and 24 is the empirically optimal value. 

\textbf{Latent World Model.} To let the latent action tokens to capture the physical dynamics between frames, we adopt an auto-regressive Transformer as the architecture of the world model with a time-causal attention mechanism. Specifically, within a single time step, the $K$ latent action tokens and the $N$ image latent tokens attend to each other bidirectionally, whereas across different time steps, attention between tokens is constrained to be causal.

\begin{table}[!htbp]
  \caption{Configuration of the Latent World Model.}
  \label{table:configuration_WM}
  \begin{center}
    \begin{small}
      \begin{sc}
        \begin{tabular}{ll}
        \toprule
        Parameter & Value \\
        \midrule
        Transformer Layers& 12 \\
        Attention heads & 8 \\
        Image token dimension & 2048 \\
        Number of image tokens per time step & 256 \\
        Action token dimension & 2048 \\ 
        Number of action tokens per time step & 3 \\
        Number of view & 2 \\
        Future video horizon & 8 \\
        \bottomrule
        \end{tabular}
      \end{sc}
    \end{small}
  \end{center}
  \vskip -0.1in
\end{table}

\textbf{Action Head.} To generate future actions conditioned on latent action embeddings, we adopt a flow matching-based Transformer architecture, DiT-B~\cite{peebles23dit}, as our action head. DiT-B is conditioned on an embodied action token sequence formed by repeating the action token 32 times, where 32 is the empirically optimal value.

\begin{table}[!htbp]
  \caption{Configuration of the Action head.}
  \label{table:configuration_AH}
  \begin{center}
    \begin{small}
      \begin{sc}
        \begin{tabular}{ll}
        \toprule
        Parameter & Value \\
        \midrule
        Transformer Layers& 16 \\
        Attention heads & 12 \\
        Token dimension & 1024 \\
        State dimension & 8 \\
        Action dimension & 7 \\ 
        Future action horizon & 7 \\
        Positional encoding & Learnable \\
        Denoising timesteps & 4 \\
        \bottomrule
        \end{tabular}
      \end{sc}
    \end{small}
  \end{center}
  \vskip -0.1in
\end{table}

\subsection{Training Details}

We preprocess all training datasets following a unified pipeline. Observation images used as inputs to the VLM are resized to 224×224. Video clips used by the world-state encoder are resized to 256×256 to be consistent with the hyperparameter settings described in the previous subsection. For models pretrained with joint-position control (e.g. $\pi_0$), we use joint-space delta positions as actions and apply min–max normalization to map them into $[0, 1]$. For models pretrained with end-effector control (e.g., \method{}), we use both end-effector delta positions and delta axis–angle representations as actions, which are likewise min–max normalized to $[0, 1]$ All gripper commands are binarized to $\{0,1\}$. To handle multi-view observations, when fewer than two camera views are available, we duplicate the world-state representation and concatenate the two copies. When more than two views are available, we select two of the available views and use them as the world-state representation.

During training, we use a batch size of 32 and train on 8 GPUs in parallel, resulting in a global batch size of 256. We adopt a cosine learning-rate schedule with linear warmup, with a peak learning rate of 1e-5 for the VLM and the latent world model, and 1e-4 for action head. For the pretraining datasets, ssv2 and droid, we jointly train the model for 50K steps. For the simulation datasets, we continue training for 30K steps starting from the last pretrained checkpoint. For the real-world datasets, we further fine-tune the model for 20K steps, also initialized from the last pretrained checkpoint.
\section{Real-world Experiments Details}
\label{appendix:real-world}

Our real-world setup consists of a Franka Research 3 robotic arm, a Robotiq 2F-85 gripper, and three Intel RealSense D435 cameras, including two third-person views and one wrist-mounted view. The collected expert demonstrations involve picking and placing grapes, apples, mangoes, and oranges from a table into a plate or a bowl. In the following experiments, each task is executed for 10 independent trials, and we report the average success rate over all trials.

For task-level OOD evaluation in real-world experiments, we consider the following tasks: (i) picking up bananas from the table and placing them into a bowl, (ii) picking up peaches from the table and placing them onto a plate, and (iii) picking up grapes from the table and placing them onto the top level of a shelf.

We observe that, for the banana-picking task, both $\pi_{0.5}$ and \method{} achieve approximately a 50\% success rate. For the peach-picking task, due to the irregular shape of peaches, the robot frequently violates the predefined safety boundaries during execution. For the final task, since shelves are not present in the training data, none of the models can successfully place the end-effector at the top level. However, compared with $\pi_{0}$ and $\pi_{0.5}$, which directly collide with the shelf, \method{} exhibits a qualitatively different behavior by approaching the target from the rear side of the shelf and lifting the end-effector to a higher position. Although the task is still not successfully completed, this behavior indicates that \method{} possesses improved task generalization ability compared with the baselines.

For object-layout–level OOD evaluation in real-world experiments, we randomly select three tasks from the training set and randomly shuffle the object layouts to assess the robustness and generalization capability of the models.

\begin{figure*}[t!]
    \centering
    \includegraphics[width=0.98\linewidth]{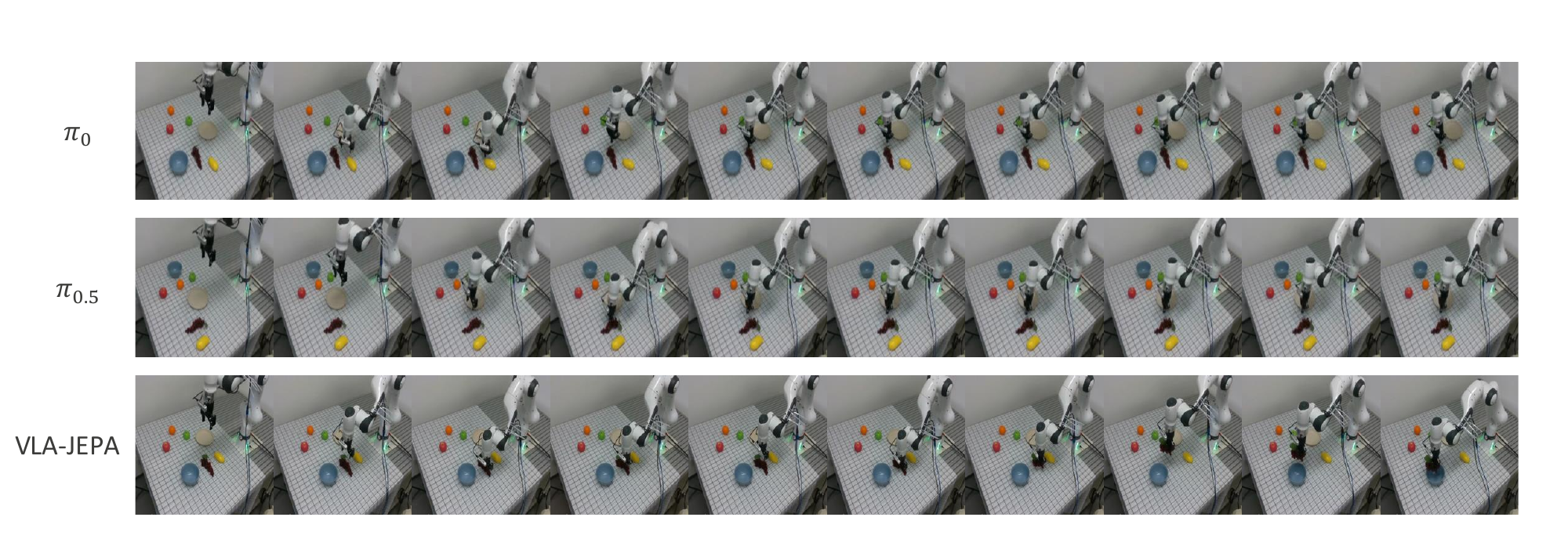}
    \caption{Comparison of three models ($\pi_0$, $\pi_{0.5}$, \method{}) under the object-layout OOD setting.}
    \label{fig:real_world_frames}
    \vspace{-10pt}
\end{figure*}

As shown in Figure~\ref{fig:real_world_frames}, we observe that both $\pi_{0}$ and $\pi_{0.5}$ fail to reattempt grasping after an unsuccessful grasp, since the training data do not contain demonstrations of repeated grasping behaviors, i.e., reopening the gripper and regrasping after a failure. Consequently, when a grasp fails, these two policies do not actively release the gripper, as they have not learned the action of opening the gripper to reattempt a grasp. In contrast, \method{}, which is pretrained on human videos and thus exposed to repeated grasping behaviors, is able to immediately open the gripper and attempt to grasp again after a failure. We argue that repeated grasping does not require learning additional physical dynamics; instead, it mainly requires the model to learn when to perform a regrasp action. Once this temporal decision is learned, the policy can internally map it to its own physical dynamics and execute the corresponding motion. This observation highlights one of the key advantages of pretraining on human videos.